\documentclass[10pt,twocolumn,letterpaper]{article}

\usepackage{cvpr}              

\usepackage{multirow}

\definecolor{cvprblue}{rgb}{0.21,0.49,0.74}
\usepackage[pagebackref,breaklinks,colorlinks,allcolors=cvprblue]{hyperref}

\def\paperID{8688} 
\def\confName{CVPR}
\def\confYear{2026}

\title{State and Scene Enhanced Prototypes for Weakly Supervised Open-Vocabulary Object Detection}

\author{Jiaying Zhou\\
National Institute of Health Data Science\\
Peking University, Beijing, China\\
{\tt\small zhoujiaying@pku.edu.cn}
\and
Qingchao Chen\\
National Institute of Health Data Science\\
Peking University, Beijing, China\\
{\tt\small qingchao.chen@pku.edu.cn}
}

\begin{document}
\maketitle
\begin{abstract}
Open-Vocabulary Object Detection (OVOD) aims to generalize object recognition to novel categories, while Weakly Supervised OVOD (WS-OVOD) extends this by combining box-level annotations with image-level labels. Despite recent progress, two critical challenges persist in this setting. First, existing semantic prototypes, even when enriched by LLMs, are static and limited, failing to capture the rich intra-class visual variations induced by different object states (e.g., a cat's pose). Second, the standard pseudo-box generation introduces a semantic mismatch between visual region proposals (which contain context) and object-centric text embeddings.
To tackle these issues, we introduce two complementary prototype enhancement strategies. To capture intra-class variations in appearance and state, we propose the State-Enhanced Semantic Prototypes (SESP), which generates state-aware textual descriptions (e.g., "a sleeping cat") to capture diverse object appearances, yielding more discriminative prototypes. Building on this, we further introduce Scene-Augmented Pseudo Prototypes (SAPP) to address the semantic mismatch. SAPP incorporates contextual semantics (e.g., "cat lying on sofa") and utilizes a soft alignment mechanism to promote contextually consistent visual-textual representations. By integrating SESP and SAPP, our method effectively enhances both the richness of semantic prototypes and the visual-textual alignment, achieving notable improvements.

\end{abstract}    
\section{Introduction}
\label{sec:intro}

\begin{figure}[htbp]
  \centering
   \includegraphics[width=0.95\linewidth]{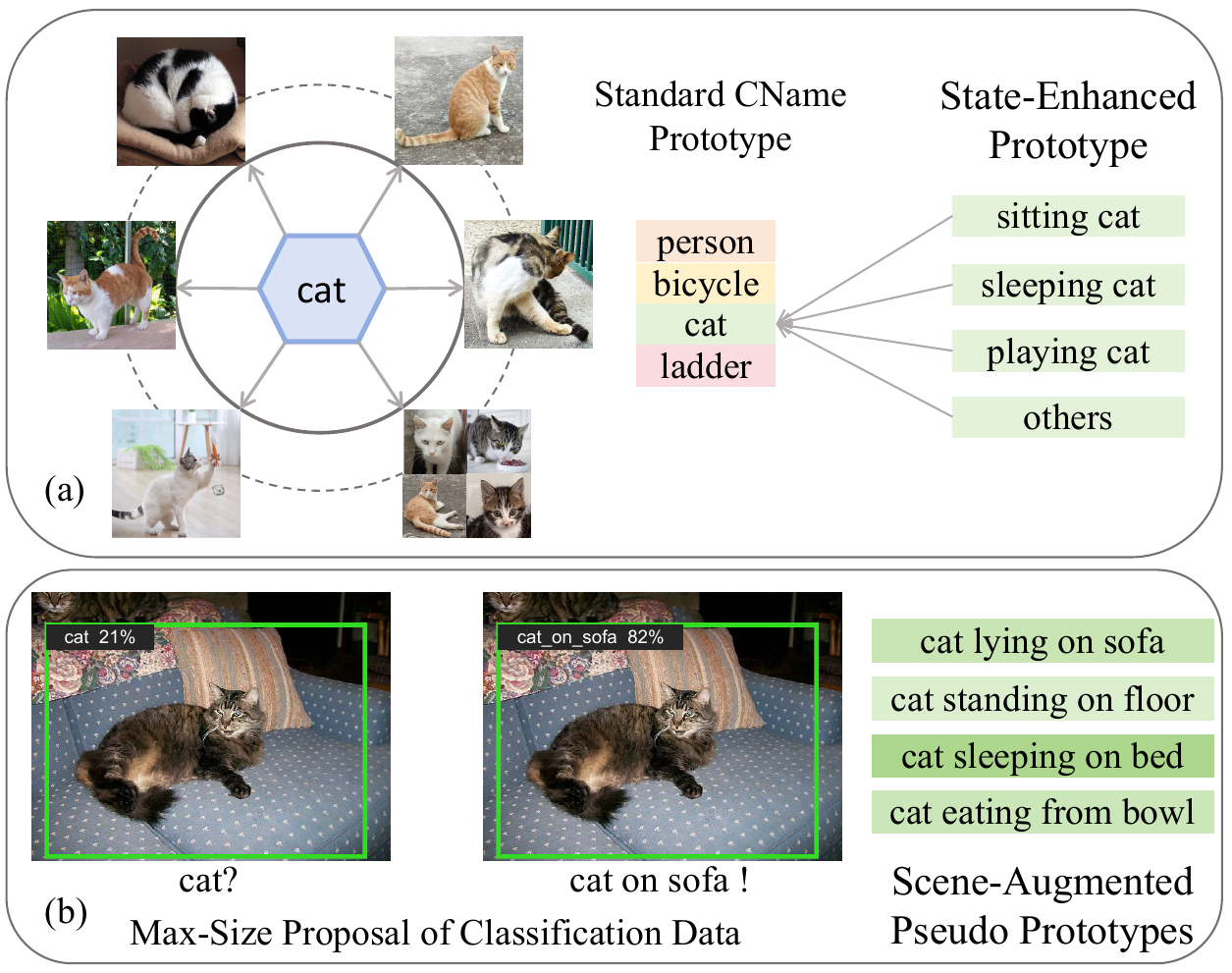}
   \caption{Semantic Prototype Augmentation for Weakly Supervised Open-Vocabulary Object Detection. (a) State-Enhanced Semantic Prototypes (SESP). Standard category names (e.g., "cat") inadequately capture intra-class visual diversity across various object states (e.g., "sleeping cat," "sitting cat"). SESP addresses this by augmenting prototypes with state-aware textual descriptions for more discriminative representations. (b) Scene-Augmented Pseudo Prototypes (SAPP). Standard pseudo-boxes (max-size proposals used in image classification data) often include substantial contextual information, leading to misalignment with class-centric prototypes. SAPP resolves this by incorporating contextual semantics to generate pseudo prototypes and using a soft alignment mechanism to bridge the visual-textual gap.}
   \label{fig:tea}
\end{figure}

Object detection aims to localize and classify all objects of interest within an image. Conventional detectors are trained under closed-set assumption, meaning they perform well only on a predefined set of categories and struggle to generalize to unseen classes, particularly in terms of classification accuracy\cite{wang2023opencorpus}. This inherent limitation has driven the development of open-vocabulary object detection (OVOD), which aims to localize and recognize novel objects beyond the pre-defined categories.

OVOD commonly exploits pre-trained vision-language models (VLMs) for recognition by substituting the learnable classification head with text embeddings of category names (e.g., “cat”, “car”). Detection is then performed by aligning region proposals with these semantic embeddings through cross-modal similarity, enabling the recognition of categories unseen during training.
To improve generalization to novel classes, Weakly Supervised Open-Vocabulary Object Detection (WS-OVOD) leverages image-level labels, typically sourced from large-scale image classification datasets containing unseen categories, as a labeling-efficient alternative to bounding box annotations. In this setting, max-size proposals derived from classification data are retained and aligned with semantic prototypes to bridge the gap between image-level supervision and object-level recognition. Despite growing interest in this paradigm, two critical challenges remain unresolved.

First, category names often convey only coarse or underspecified semantic information. Previous methods have attempted to enrich semantic prototypes by prompting large language models (LLMs) to generate generic textual descriptions. However, such descriptions remain static and fail to reflect the rich intra-class visual variations induced by object states and contextual factors. For example, as illustrated in \cref{fig:tea}(a), a cat may appear in various poses—such as sitting, sleeping, or grooming—each corresponding to distinct visual configurations.
To overcome this limitation, we propose the \textit{State-Enhanced Semantic Prototypes} (SESP), in which category-specific textual descriptions are generated by prompting a language model with diverse object states (e.g., “a sleeping cat is curled or stretched out, typically appearing relaxed and content”, or analogous prompts for states such as sitting, running, etc.). These state-aware descriptions are incorporated into the semantic prototypes alongside standard (state-agnostic) category descriptions. By explicitly modeling intra-class variations induced by object states, SESP better reflects the visual diversity of object instances, leading to more discriminative and informative semantic prototypes.

Second, the common pseudo-box strategy in WS-OVOD, which selects the max-size region proposal as the pseudo-box, introduces a persistent semantic mismatch between visual and textual representations. As shown in \cref{fig:tea}(b), although these max-size proposals typically cover the target object, they also encompass surrounding context that is absent from the object-centric semantic prototypes. Typical prototypes, even when enhanced, focus solely on the object itself, neglecting the implicit contextual information present in the visual features. To mitigate this issue, we extend the original semantic prototypes by introducing a group of \textit{scene-augmented pseudo prototypes}(SAPP). These pseudo prototypes incorporate contextual semantics while remaining anchored around the target object(e.g., “cat lying on sofa”). During training, a soft alignment mechanism is applied to associate the scene-augmented pseudo prototypes with corresponding region proposals, promoting more contextually consistent visual-textual representations and thereby reducing the semantic gap.

By integrating state-enhanced semantic prototypes (SESP) with scene-augmented pseudo prototypes (SAPP), our method effectively addresses two key challenges in weakly supervised open-vocabulary object detection: intra-class variation due to object states and semantic mismatch from contextual discrepancy, leading to consistent performance gains on both base and novel categories. The main contributions of this work are summarized as follows:

\begin{itemize}
    \item We design state-enhanced semantic prototypes that explicitly incorporate descriptions of different object states, enabling richer and more flexible category representations.
    \item  We propose scene-augmented pseudo prototypes to utilize contextual scene information, enhancing the semantic expressiveness of detected objects.
    \item Comprehensive experiments on WS-OVOD benchmarks demonstrate that our method surpasses existing approaches, validating its effectiveness.
\end{itemize}






\section{Related Work}

\subsection{Open-Vocabulary Object Detection}
Open-vocabulary object detection aims to enable the model to generalize to arbitrary categories through cross-modal alignment, detecting objects unseen during training. This task was first proposed by OVR-CNN~\cite{OVR-CNN}, which adapted Faster R-CNN into an open-vocabulary framework by leveraging image-caption datasets. Existing solutions broadly fall into four categories: (1) fine-tuning vision-language models (VLMs), (2) knowledge distillation, (3) prompt learning, and (4) pseudo-labeling under weak supervision.

Fine-tuning-based approaches adapt pretrained VLMs by constructing region-level data pairs and fine-tuning VLMs using contrastive objectives to achieve region-level alignment \cite{rovit,lin2022learning,zhong2022regionclip,liu2024groundingdino}. Although these methods achieve competitive performance, they require extensive labeled data and substantial computational resources. Moreover, the domain gap between pretraining corpora and detection scenarios can hinder effective transfer.

Knowledge distillation paradigms exploit the zero-shot generalization ability of large VLMs to transfer knowledge to object detectors. Techniques such as ViLD \cite{gu2021open} align detector region features with CLIP embeddings, while methods like BARON \cite{wu2023aligning} aggregate contextually related proposals to enhance alignment. Hierarchical distillation frameworks \cite{ma2022open,oadp,hpovd} further integrate multi-level knowledge transfer, encompassing instance, region, and global semantics. Other distillation variants, e.g., DK-DETR \cite{dkdetr} and CAKE \cite{ma2025cake}, refine semantic and relational knowledge transfer, emphasizing category-specific nuances. Despite their efficiency, these methods often rely on fixed VLM features and may struggle to adapt to fine-grained visual distinctions in complex scenes.

Prompt-learning-based methods aim to enrich semantic category representations beyond simple name embeddings by generating descriptive textual features. For instance, MM-OVOD \cite{kaul2023multi} leverages large language models (LLMs) to create multiple descriptive prompts per category, while DVDet \cite{jin2024llms} produces fine-grained descriptors to integrate into embeddings. Hierarchical semantics are exploited by SHiNe \cite{liu2024shine} to provide structured prompts. Additionally, learnable prompts \cite{du2022learning,feng2022promptdet,ren2023prompt,zhao2024scene} and visual prompts \cite{kaul2023multi,zhao2024scene} have been proposed to further enhance semantic prototypes. However, existing semantic prototypes fail to model intra-class variations caused by diverse object states and contextual factors. To address this limitation, we propose \textit{state-enhanced semantic prototypes} (SESP), which incorporate \textit{state-specific} textual descriptions to better capture intra-class diversity.

Pseudo-labeling approaches tackle open-vocabulary object detection by leveraging weak supervision, typically from image classification datasets. A common strategy in pseudo-labeling selects the max-size proposals generated by the detector as pseudo-boxes and aligns them with prototypes~\cite{zhou2022detecting,kaul2023multi,jeong2024proxydet,huang2024open,jin2024llms}. However, these max-size proposals often include substantial contextual and background information, while the original category semantic prototypes focus only on the target objects, leading to a \textit{semantic mismatch} between context-rich visual regions and object-centric semantic prototypes. Recent efforts also explore iterative refinement of pseudo-labels or uncertainty-aware selection to improve reliability, yet the fundamental representation misalignment remains largely unaddressed. To mitigate this gap, we propose \textit{scene-augmented pseudo prototypes} (SAPP) to utilize contextual scene information, thereby improving the alignment of max-size proposals with category semantics.

\subsection{Weakly-Supervised Object Detection}
Conventional object detection methods heavily depend on large-scale, precisely annotated bounding box data, which is often costly and labor-intensive to obtain. In contrast, weakly supervised object detection (WSOD) aims to leverage only image-level labels for detector training \cite{shao2022deep}, thereby significantly reducing annotation overhead. A prominent line of work in WSOD employs multiple instance learning (MIL) frameworks to localize discriminative object regions within images without requiring bounding boxes \cite{bilen2016weakly,ren2020instance,tang2017multiple,li2019weakly,wan2019c}. Complementary techniques have utilized Class Activation Mapping (CAM) and its variants to tackle object localization challenges under weak supervision \cite{diba2017weakly,zhang2018adversarial}. Additionally, some approaches adopt hybrid supervision strategies, jointly exploiting image-level and limited box-level annotations to improve detection performance. These methods often model image-level label supervision as a proposal classification task, either selecting a single proposal that best matches the image label \cite{redmon2017yolo9000,ramanathan2020dlwl} or conducting many-to-many matching between multiple proposals and labels to better capture complex object distributions \cite{lin2022learning}. While traditional weakly supervised object detection (WSOD) focuses on seen categories, its extension to open-vocabulary settings, in which novel categories appear at test time, poses additional challenges in semantic generalization and cross-modal alignment. Our work addresses these challenges in the context of weakly supervised open-vocabulary object detection (WS-OVOD). 
\section{Method}

\begin{figure*}[htbp]
  \centering
   \includegraphics[width=0.98\textwidth]{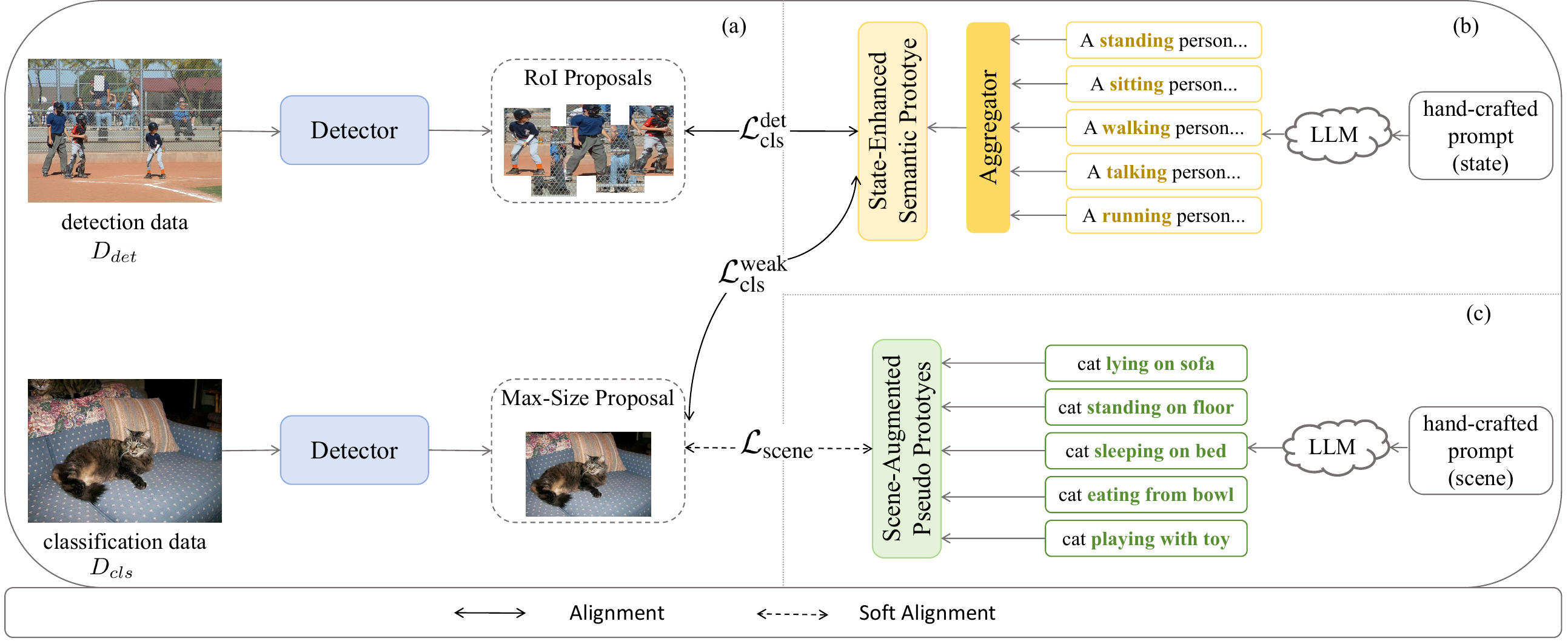}
   \caption{Overview of our proposed method. (a) For detection data $D_{det}$, RoI features are aligned to state-enhanced semantic prototypes $\textit{\textbf{p}}_c$. For classification data $D_{cls}$, only the max-size proposal is retained and its feature is aligned with $\textit{\textbf{p}}_c$ for weakly supervised classification and softly aligned with scene-augmented pseudo prototypes $\textit{\textbf{w}}_{scene,c}$ to capture contextual semantics. (b) State-enhanced semantic prototypes are generated by prompting a large language model (LLM) with state-specific templates, followed by aggregation across diverse state descriptions. (c) Scene-augmented pseudo prototypes are derived by prompting the LLM with context-aware templates, capturing object-scene interactions to enrich representation. Note that the RPN loss $\mathcal{L}_{\text{rpn}}$ and the bounding box regression loss $\mathcal{L}_{\text{reg}}$, which constitute the standard detection losses trained only on $D_{\text{det}}$, are omitted for clarity.}
   \label{fig:main}
\end{figure*}

In this work, we focus on weakly supervised open-vocabulary object detection (WS-OVOD), addressing two major challenges in existing approaches: (1) the limited capacity of static category names to effectively capture rich intra-class visual variations, and (2) the inherent semantic gap between object-centric category embeddings and context-rich, weakly supervised visual proposals. To overcome these issues, we introduce two components: state-enhanced semantic prototypes (SESP), which enrich category representations by incorporating object state information; and scene-augmented pseudo prototypes (SAPP), designed to bridge contextual discrepancies within weak supervision. We formally define the task in \cref{subsec:task}, followed by a detailed description of SESP and SAPP in \cref{subsec:state} and \cref{subsec:scene}, respectively.

\subsection{Task Definition}
\label{subsec:task}
Weakly Supervised Open-Vocabulary Object Detection (WS-OVOD) aims to localize and recognize objects from an open set of categories by leveraging both fully annotated detection data and weakly annotated image-level classification data.
The training dataset comprises two parts: 

(1) a detection dataset $D_{det} = \{(\textit{\textbf{x}}, \textit{\textbf{y}}_{det})\}$, where $\textit{\textbf{x}}$ denotes an input image and $\textit{\textbf{y}}_{det} = \{(\textit{\textbf{b}}_i, c_i)\}$ represents the set of bounding box coordinates $\textit{\textbf{b}}_i$ and corresponding class labels $c_i$ for objects in $\textit{\textbf{x}}$; 

(2) a classification dataset $D_{cls} = \{(\textit{\textbf{x}}, y_{cls})\}$, where $\textit{\textbf{x}}$ denotes an image and $y_{cls}$ is the image-level class label. Notably, $D_{cls}$ can contain categories not present in $D_{det}$.

During training, the object detector is jointly optimized on both datasets with the following combined loss:

\begin{equation}
    \mathcal{L} = \mathcal{L}_{det} + \mathcal{L}_{weak},
\end{equation}

where $\mathcal{L}_{\text{det}} = \mathcal{L}_{\text{rpn}} + \mathcal{L}_{\text{reg}} + \mathcal{L}_{\text{cls}}^{\text{det}}$ is the standard detection loss, and $\mathcal{L}_{\text{weak}} = \mathcal{L}_{\text{cls}}^{\text{weak}}$ is the weakly supervised classification loss computed using image-level labels and pseudo-boxes (taken as the max-size proposal by area).

During training, for images from $D_{det}$, the detector generates region proposals via a Region Proposal Network (RPN), extracts RoI features $\textit{\textbf{w}}_r$, and predicts refined bounding boxes and classification scores by computing cosine similarity between $\textit{\textbf{w}}_r$ and $\textit{\textbf{p}}_c$ for each category $c$, where $\textit{\textbf{p}}_c$ denotes the state-enhanced semantic prototype produced by our SESP module. For images from $D_{cls}$, only the max-size proposal is retained as the pseudo-box. 
Its RoI feature $\textit{\textbf{w}}_{mr}$ is used for classification via alignment with $\textit{\textbf{p}}_c$, and simultaneously undergoes soft alignment with the scene-augmented pseudo prototype $\textit{\textbf{w}}_{scene,c}$ to promote contextual consistency, as shown in \cref{fig:main}. 
The visual backbone and RoI classification head are trained on both datasets, while the RPN and bounding box regressor are updated only with $D_{det}$, as image-level labels lack precise localization supervision.

\subsection{State-Enhanced Semantic Prototypes}
\label{subsec:state}

In open-vocabulary object detection (OVOD), semantic prototypes are typically represented by embeddings of category names. However, such prototypes often lack sufficient semantic richness and tend to be ambiguous. A key limitation is that objects within the same category can exhibit significant variations in visual appearance depending on their different states (e.g., a “sleeping cat” vs. a “running cat”). These intra-class variations are crucial and can lead to substantial differences in how objects manifest visually. While prior works have attempted to enrich semantic prototypes using category descriptions or attribute information, these approaches generally provide broad and generic descriptions that fail to capture the diverse state-dependent characteristics within each class. In this work,  we propose \textbf{State-Enhanced Semantic Prototypes (SESP)}, which leverage descriptive information about different states or forms of objects within the same category. By incorporating these state-specific descriptions, our approach captures finer-grained semantic nuances and better models the intra-class appearance variations. In the following, we will detail the process of generating state descriptions and the construction of state-enhanced semantic prototypes.

\noindent\textbf{State Description Generation.} 
We automatically generate descriptive state prompts for each category by querying a large language model (LLM) with the following template:

\begin{quote}
\texttt{What are the common states or forms of \textit{C} (category name)? For each common state of \textit{C}, provide a one-sentence description of its visual appearance.}
\end{quote}

Using this template, the LLM produces visual descriptions for the top-$K$ most common states, denoted $\{ state_c^k \}_{k=1}^K$, for category $c$. Each description explicitly includes the state and category name (e.g., “a sleeping cat”).
These textual descriptions are then encoded into embeddings via a VLM text encoder $f_t(\cdot)$:
\begin{equation}
    \textit{\textbf{w}}_{state,c}^k = f_t(state_c^k), \quad k=1,\dots,K.
\end{equation}

\noindent\textbf{Generic Category Description.} 
While state-specific descriptions capture within-class variation, they primarily encode state-level semantics and may overlook shared, category-level characteristics. To complement this, we also generate a generic category description by prompting the LLM as follows:

\begin{quote}
\texttt{What does \textit{C} (category name) generally look like? Provide a one-sentence generic description of its appearance.}
\end{quote}

The resulting description $des_c$ is encoded similarly:

\begin{equation}
    \textit{\textbf{w}}_{des,c} = f_t(des_c).
\end{equation}

\noindent\textbf{Prototype Aggregation.} Finally, the state embeddings $\{\mathbf{w}_{state,c}^k\}_{k=1}^K$ and the generic description embedding $\mathbf{w}_{des,c}$ are aggregated to form the final semantic prototype $\mathbf{p}_c$:

\begin{equation}    
\textit{\textbf{p}}_c = \operatorname{Aggregator}\big(\{\textit{\textbf{w}}_{state,c}^k\}_{k=1}^K, \textit{\textbf{w}}_{des,c}\big).
\end{equation}

In our implementation, we adapt a simple yet effective mean aggregation:
\begin{equation}    
\textit{\textbf{p}}_c = \frac{1}{K+1} \left( \textit{\textbf{w}}_{des,c} + \sum_{k=1}^K \textit{\textbf{w}}_{state,c}^k \right).
\end{equation}

where $\mathbf{p}_c$ denotes the state-enhanced semantic prototype for class $c$. This unified prototype $\mathbf{p}_c$ captures both the general and diverse state-specific semantic cues, improving representation fidelity for open-vocabulary object detection.

\subsection{Scene-Augmented Pseudo Prototypes}
\label{subsec:scene}
Open-vocabulary object detection faces the fundamental challenge of lacking bounding box annotations for novel classes. To address this, weakly supervised methods commonly utilize image classification datasets as auxiliary training data. A typical approach involves applying the detector to extract region proposals and assigning the max-size proposal as the pseudo bounding box, which is subsequently used for classification. 

While the max-size proposal often captures the target object, it also includes substantial background and contextual content. This leads to a semantic misalignment between the context-rich proposal features and the object-centric semantic prototypes, which describe only the target category without accounting for scene context. Such misalignment can degrade detection performance, particularly for novel classes. To bridge this gap, we propose \textbf{scene-augmented pseudo prototypes(SAPP)} that extend semantic prototypes with contextual information extracted from scenes. Specifically, we first generate scene-aware phrases for each category by querying a large language model (LLM) with the following template:
\begin{quote}
\texttt{In which contexts is \textit{C} most commonly found? Please output phrases in the form of '\textit{C} + context'.}
\end{quote}

This yields $L$ distinct scene-augmented phrases $\{scene_c^l\}_{l=1}^L$ for class $c$, where each phrase provides category-centered contextual information. We encode these phrases into embeddings using the text encoder $f_t$ introduced above:
\begin{equation}
    \textit{\textbf{w}}_{scene,c}^l = f_t(scene_c^l), \quad l=1,\dots,L.
\end{equation}

where $\textit{\textbf{w}}_{scene,c}^l$ denotes the $l$-th scene-augmented pseudo prototype for class $c$. Collectively, these form an expanded pseudo prototype set $\mathcal{W} = \{ \textit{\textbf{w}}_{scene, c}^{l} \mid c \in \mathcal{C},\, l = 1,\dots,L \}$ that captures both categorical and contextual semantics.

During training, for each image from the weakly labeled classification dataset, we retain the max-size region proposal (by area) as a pseudo-box. Let $\mathbf{w}_{mr,b}$ denote the visual embedding of $b$-th max-size proposal in a batch. We compute its cosine similarity with all scene-augmented pseudo prototypes as:
\begin{equation}
    s_{b, c, l} = cosine(\textit{\textbf{w}}_{mr,b}, \textit{\textbf{w}}_{scene,c}^{l}), \quad \forall c \in \mathcal{C},\, l = 1,\dots,L.
\end{equation}

For the ground-truth image-level label $c^\ast$, we treat the $L$ corresponding pseudo prototypes $\{ \textit{\textbf{w}}_{scene, c^\ast}^{l} \}_{l=1}^L$ as candidate positive labels. A pseudo-label $y_{b, c, l} \in \{0, 1\}$ is assigned based on similarity threshold $\tau$:
\begin{equation}
    y_{b, c, l} =
    \begin{cases}
        1, & \text{if } c = c^\ast \text{ and } s_{b, c, l} \geq \tau, \\
        0, & \text{otherwise}.
    \end{cases}
\end{equation}
The associated label weight is set to the sigmoid of the similarity score: $w_{b, c, l} = \sigma(s_{b, c, l})$, which serves as a soft confidence measure.

Finally, we employ a confidence-weighted multi-label binary cross-entropy loss:
\begin{equation}
    \mathcal{L}_{\text{scene}} = -\frac{1}{B} \sum_{b=1}^B \sum_{c \in \mathcal{C}} \sum_{l=1}^L w_{b,c,l} \, \ell_{\text{bce}}\big(y_{b,c,l}, s_{b,c,l}\big),
\end{equation}
where $\ell_{\text{bce}}(y, s) = y \log \sigma(s) + (1 - y) \log (1 - \sigma(s))$ and $\sigma(\cdot)$ denotes the sigmoid function.

\subsection{Overall Objective}
\label{subsec:objective}
We augment the standard weakly supervised open-vocabulary object detection (WS-OVOD) objective with the proposed scene-augmented pseudo prototype alignment loss to better exploit contextual cues during training. The full training loss is given by:
\begin{equation}
    \mathcal{L} = \mathcal{L}_{\text{det}} + \mathcal{L}_{\text{weak}} + \lambda \, \mathcal{L}_{\text{scene}},
\end{equation}
where $\mathcal{L}_{\text{scene}}$, defined in Section~\ref{subsec:scene}, aligns max-size proposals with scene-augmented pseudo prototypes to encourage context-aware semantic matching. The hyperparameter $\lambda$ controls its contribution and is set to 0.1 in all experiments.

\section{Experiment}

\subsection{Experimental Setups}
\noindent\textbf{Datasets}
We evaluate our method on two standard open-vocabulary object detection benchmarks: COCO~\cite{lin2014microsoft} and LVIS~\cite{gupta2019lvis}, following the conventional protocols in~\cite{gu2021open,zhou2022detecting}, denoted as OV-COCO and OV-LVIS, respectively.

In OV-COCO, the dataset is split into 48 base classes and 17 novel classes. We report AP$_{50}$ (average precision at IoU=0.5) for all, base (AP$^b_{50}$), and novel (AP$^n_{50}$) categories.

For OV-LVIS, the 1,203 categories are partitioned into frequent, common, and rare subsets. Following~\cite{gu2021open,zhou2022detecting}, we treat the 337 rare categories as novel classes and the remaining 866 (frequent + common) as base classes. Detection performance is evaluated using  standard AP@[0.5:0.95]. We report results for rare (AP$_r$), common (AP$_c$), frequent (AP$_f$), and all categories (AP), with AP$_r$ serving as the primary metric for open-vocabulary performance. 

To support weakly supervised open-vocabulary detection, we further employ large-scale image classification datasets, ImageNet-LVIS~\cite{deng2009imagenet} and Conceptual Captions~\cite{sharma2018conceptual}, as auxiliary sources of semantic supervision. ImageNet-LVIS is the subset of ImageNet-21K that contains only the categories that intersect with those of LVIS. Conceptual Captions is an image captioning dataset with 3M images. Following Detic~\cite{zhou2022detecting}, we extract image labels via exact text matching and keep only images whose captions mention at least one LVIS category, providing weak supervision for training.

\noindent\textbf{Implementation details}
We adapt Detic~\cite{zhou2022detecting} as our baseline framework. For OV-COCO, we employ Faster R-CNN~\cite{ren2016fasterrcnn} with a ResNet-50~\cite{he2016deepresnet} backbone to align with prior open-vocabulary baselines; for OV-LVIS, we follow Detic and utilize CenterNet2~\cite{zhou2021probabilisticcenternet2}. 
For each class, we generate multiple state- and scene-aware textual descriptions using GPT-4o, and form the final semantic prototype by averaging their CLIP text embeddings. Additional implementation details are provided in the appendix.

\subsection{Comparison with State-of-the-Arts}
\noindent\textbf{Results on OV-COCO} 
In \cref{tab:ovcoco_results}, we compare our method with state-of-the-art approaches on the OV-COCO benchmark under the weakly supervised open-vocabulary setting. Our approach achieves an AP$^n_{50}$ of 31.3 and an overall AP$_{50}$ of 47.5, outperforming Detic~\cite{zhou2022detecting} (under the same experimental setting) by 3.5 and 2.5 points, respectively.
These results demonstrate the effectiveness of our proposed state-enhanced semantic prototypes and scene-augmented pseudo prototypes. Notably, the improvement on novel classes (3.5 points) exceeds that on base classes (1.5 points), indicating that our method is particularly effective for novel category detection under weak supervision.

\begin{table}[t]
\centering
\caption{Performance comparison on the OV-COCO benchmark under the weakly supervised open-vocabulary object detection (WS-OVOD) setting. All methods use a ResNet-50 backbone unless otherwise specified.}
\label{tab:ovcoco_results}
\setlength{\tabcolsep}{12pt}
\begin{tabular}{cccc}
\toprule
Method &   AP$^n_{50}$  & AP$^b_{50}$ & AP$_{50}$ \\
\midrule
Box-Supervised & 1.3 & 52.8 & 39.3 \\
\midrule
Detic~\cite{zhou2022detecting} &  27.8 & 47.1 & 45.0 \\
DCS~\cite{zhang2024opendcs} & 30.3 & 52.5 & 46.7 \\
ProxyDet~\cite{jeong2024proxydet} &  30.4 & 52.6 & 46.8 \\
SIC-CADS~\cite{fang2024simple} & 31.0 & 52.4 & 46.8 \\
DVDet\cite{jin2024llms} &  29.5 & \textbf{53.6} & 47.3 \\
CODet~\cite{li2025benefit} & 29.8 & 52.9 & 46.8 \\
HMKM~\cite{ma2025hierarchical} & 29.5 & 50.8 & 45.3 \\
\midrule
Ours &  \textbf{31.3} & 53.3 & \textbf{47.5} \\
\bottomrule
\end{tabular}
\end{table}

\noindent\textbf{Results on OV-LVIS} 
In \cref{tab:ovlvis_results}, we compare our method with state-of-the-art approaches on the OV-LVIS benchmark. Our approach consistently outperforms Detic~\cite{zhou2022detecting} under the same weakly supervised open-vocabulary setting and backbone architecture. Specifically, with a ResNet-50 backbone, our method achieves an AP$_r$ of 28.0, surpassing Detic by 3.1 points. When equipped with a Swin Transformer (base) backbone, our method achieves 36.7 AP$_r$, surpassing Detic by 2.9 points. These consistent gains demonstrate that our prototype enhancement, integrating state-enhanced semantic prototypes and scene-augmented pseudo prototypes, effectively bridges the semantic gap in weakly supervised open-vocabulary detection, regardless of backbone architecture.

\begin{table*}[t]
\centering
\caption{Performance on the OV-LVIS benchmark. All comparisons are conducted under the same experimental setting. \textit{RN50} and \textit{Swin-B} denote models using ResNet-50 and Swin Transformer (base) backbones, respectively. \textit{LVIS-base} denotes training on LVIS while utilizing supervision from base classes only. \textit{IN-L} refers to the ImageNet--LVIS dataset, and \textit{CC} denotes the Conceptual Captions dataset, both used as sources of image-level weak supervision.}
\label{tab:ovlvis_results}
\setlength{\tabcolsep}{12pt}
\begin{tabular}{ccccccc}
\toprule
Method & Backbone & Supervision & AP$_r$ & AP$_c$ & AP$_f$ & AP \\
\midrule
Box-Supervised & RN50 & LVIS-base & 16.3 & 31.0 & 35.4 & 30.0 \\
\midrule
Detic~\cite{zhou2022detecting} & RN50 &  LVIS-base + IN-L & 24.9 & 33.0 & 35.4 & 32.4 \\
MM-OVOD\cite{kaul2023multi} & RN50 & LVIS-base + IN-L & 27.1 & - & - & 33.1 \\
ProxyDet~\cite{jeong2024proxydet} &  RN50 & LVIS-base + IN-L & 26.2 & \textbf{33.1} & 35.6 & 32.5 \\
SIC-CADS~\cite{fang2024simple} & RN50 & LVIS-base + IN-L & 26.5 & 33.0 & 35.6 & 32.9 \\
DCS~\cite{zhang2024opendcs} & RN50 & LVIS-base + IN-L & 26.9 & - & - & \textbf{33.3} \\
Ours & RN50 & LVIS-base + IN-L & \textbf{28.0} & 32.9 & \textbf{35.8} & \textbf{33.3} \\
\midrule
Box-Supervised & SwinB & LVIS-base & 21.9 & 40.5 & 43.3 & 38.4 \\
\midrule
Detic~\cite{zhou2022detecting}& SwinB & LVIS-base+CC & 23.9 & 40.2 & 42.8 & 38.4 \\
DVDet\cite{jin2024llms} & SwinB & LVIS-base+CC & 25.2 & \textbf{41.4} & 44.6 & \textbf{40.4} \\
CODet~\cite{li2025benefit} & SwinB & LVIS-base+CC & 25.6 & 41.1 & 44.3 & 40.2 \\
Ours & SwinB & LVIS-base+CC & \textbf{26.2} & 41.1 & \textbf{44.7} & \textbf{40.4} \\
\midrule
Detic~\cite{zhou2022detecting} & SwinB & LVIS-base + IN-L & 33.8 & 41.3 & 42.9 & 40.7 \\
HMKM~\cite{ma2025hierarchical} & SwinB & LVIS-base + IN-L & 35.4 & 41.0 & 42.7 & 40.7 \\
Ours & SwinB & LVIS-base + IN-L & \textbf{36.7} & \textbf{42.1} & \textbf{43.8} & \textbf{41.7} \\
\bottomrule
\end{tabular}
\end{table*}

\subsection{Ablation Studies}

\noindent\textbf{Ablation Studies on SESP and SAPP}
To systematically evaluate the effectiveness of our proposed components, state-enhanced semantic prototypes (SESP) and scene-augmented pseudo prototypes (SAPP), we conduct ablation studies on the OV-COCO benchmark. As shown in \cref{tab:ablation}, each component individually improves performance over the baseline, with SESP contributing +1.8 AP$^n_{50}$ points and SAPP adding +2.1 AP$^n_{50}$ points. When combined, they yield a further gain of +3.5 AP$^n_{50}$ points, demonstrating their complementary roles: SESP enriches category semantics with state diversity, while SAPP aligns visual proposals with contextualized language representations. This complementary effect validates our design principle: capturing both intra-class variation through states and inter-object context through scenes is crucial for robust weakly supervised open-vocabulary detection.

\begin{table}[t]
    \centering
    \caption{Ablation study on the OV-COCO benchmark evaluating the contributions of two key components: State-Enhanced Semantic Prototypes (SESP) and Scene-Augmented Pseudo-Labels (SAPL).}
    \label{tab:ablation}
    \setlength{\tabcolsep}{9pt} 
    \begin{tabular}{cccc}
    \toprule
    Method & AP$_{50}^n$ & AP$_{50}^b$ & AP$_50$ \\
    \midrule
    Baseline  & 27.8 & 47.1 & 45.0 \\
    + SASP & 30.3 & 52.5 &  46.7 \\
    + SAPL  &  29.2 & 52.1  & 46.1  \\
    Full (Ours) &  \textbf{31.3} & \textbf{53.3}  &  \textbf{47.5} \\
    \bottomrule
    \end{tabular}
\end{table}

\begin{table}[t]
    \centering
    \caption{Ablation study on the number of state descriptions used to enhance semantic prototypes, evaluated by AP$^n_{50}$ on the OV-COCO benchmark.}
    \label{tab:ablation_k}
    \setlength{\tabcolsep}{9pt} 
    \begin{tabular}{ccccc}
    \toprule
    $K$ & 3 & 5 & 7  & 9  \\
    \midrule
    AP$_{50}^n$ & 30.0 & 31.3 & 31.5 & 30.6  \\
    \bottomrule
    \end{tabular}
\end{table}

\noindent\textbf{Ablation Studies on $K$ and $L$}
We conduct ablation studies on OV-COCO to analyze the impact of two key hyperparameters: $K$, the number of state descriptions used to enhance semantic prototypes, and $L$, the number of scene phrases used to construct scene-augmented pseudo prototypes (SAPP) during training. As shown in \cref{tab:ablation_k} and \cref{tab:ablation_l}, performance peaks at $K=7$ and $L=5$, respectively. Using larger values leads to slight degradation, likely because redundant or noisy descriptions harm representation quality. This suggests that moderate diversity in both state and scene prompts achieves the best balance for generalization.
Although $K=7$ yields the highest performance, the gain over $K=5$ is marginal, while larger $K$ risks introducing noisy or redundant state descriptions. Since $L=5$ is optimal, we adopt $K=5$ and $L=5$ as our default setting for a balanced and efficient configuration.

\begin{table}[tbp]
    \centering
    \caption{Ablation study on the number of scene phrases used to augment prototypes (i.e., the number of pseudo prototypes for a single class), evaluated by AP$^n_{50}$ on the OV-COCO benchmark.}
    \label{tab:ablation_l}
    \setlength{\tabcolsep}{9pt} 
    \begin{tabular}{ccccc}
    \toprule
    $L$ & 3 & 5 & 7  & 9  \\
    \midrule
    AP$_{50}^n$ & 30.5 & 31.3 & 29.8 & 30.0 \\
    \bottomrule
    \end{tabular}
\end{table}

\noindent\textbf{Ablation Study on Aggregation Strategies for SESP}
In our method, the state-enhanced semantic prototype (SESP) for category $c$ is constructed by averaging the embeddings of the generic class description and multiple state-specific descriptions. To justify this design choice, we conduct an ablation study on the aggregation strategy. We compare the following four variants:

\smallskip\noindent\textbf{Mean (Ours)}: Compute the arithmetic mean of all embeddings (1 generic + $K$ state-specific).

\smallskip\noindent\textbf{Median}: Replace the mean with the element-wise median across all embeddings.

\smallskip\noindent\textbf{Two-stage mean}: First average the $K$ state-specific embeddings, then average the result with the generic embedding (i.e., equal weight to generic and state-aggregated representations).

\smallskip\noindent\textbf{Similarity-weighted}: Assign each embedding a weight proportional to its cosine similarity with the generic class embedding, then compute a weighted average.

\begin{table}[t]
\centering
\caption{Ablation on aggregation strategies for constructing state-enhanced semantic prototypes (SESP). Results are reported on OV-COCO. }
\label{tab:aggregation_ablation}
\setlength{\tabcolsep}{10pt}
\begin{tabular}{cccc}
\toprule
Aggregation Strategy & AP$_{50}^n$ & AP$_{50}^b$ & AP$_{50}$ \\
\midrule
Median &  31.2  & 53.1 & 47.4 \\
Two-stage mean & 30.4 & 52.7  & 46.9  \\
Similarity-weighted &  31.0 &  53.1 &  47.3  \\
\textbf{Mean (Ours)} & \textbf{31.3} & \textbf{53.3} & \textbf{47.5} \\
\bottomrule
\end{tabular}
\end{table}

Results in \cref{tab:aggregation_ablation} show that the choice of aggregation strategy has a relatively minor impact on performance. The standard mean pooling (Ours) achieves the best results across all metrics, but the median and similarity-weighted variants perform comparably, with differences of less than 0.1–0.2 AP$_{50}^n$ on novel categories. This suggests that the rich semantic content of the LLM-generated descriptions is robust to mild variations in aggregation. In contrast, the two-stage mean—by enforcing equal weighting between the generic and state-aggregated representations—slightly underperforms, indicating that direct averaging of all descriptions better preserves semantic diversity.

We also explored learnable aggregation mechanisms, such as (i) concatenating all embeddings and projecting them via an MLP, and (ii) computing attention weights conditioned on the generic class embedding. However, these approaches led to unstable training and inconsistent convergence, likely due to the limited supervision in the weakly supervised open-vocabulary setting. Given the marginal gains of alternative fixed-weight schemes and the instability of learnable variants, we adopt the simple and stable mean pooling in our final design.

\begin{table}[t]
\centering
\caption{Zero-shot cross-dataset object detection on the Objects365 dataset, with models trained on OV-LVIS.}
\label{tab:objects365}
\setlength{\tabcolsep}{8pt}
\begin{tabular}{ccccc}
\toprule
Methods  & AP & AP$_{50}$ & AP$_{r}$ \\
\midrule
Detic~\cite{zhou2022detecting}  & 15.6 & 22.2 & 12.4 \\
MM-OVOD-Text~\cite{kaul2023multi} & 16.6 & 23.1 & 13.1 \\
Ours & \textbf{16.9} & \textbf{23.5} & \textbf{13.7} \\
\bottomrule
\end{tabular}
\end{table}

\begin{figure*}[htbp]
  \centering
   \includegraphics[width=0.98\textwidth]{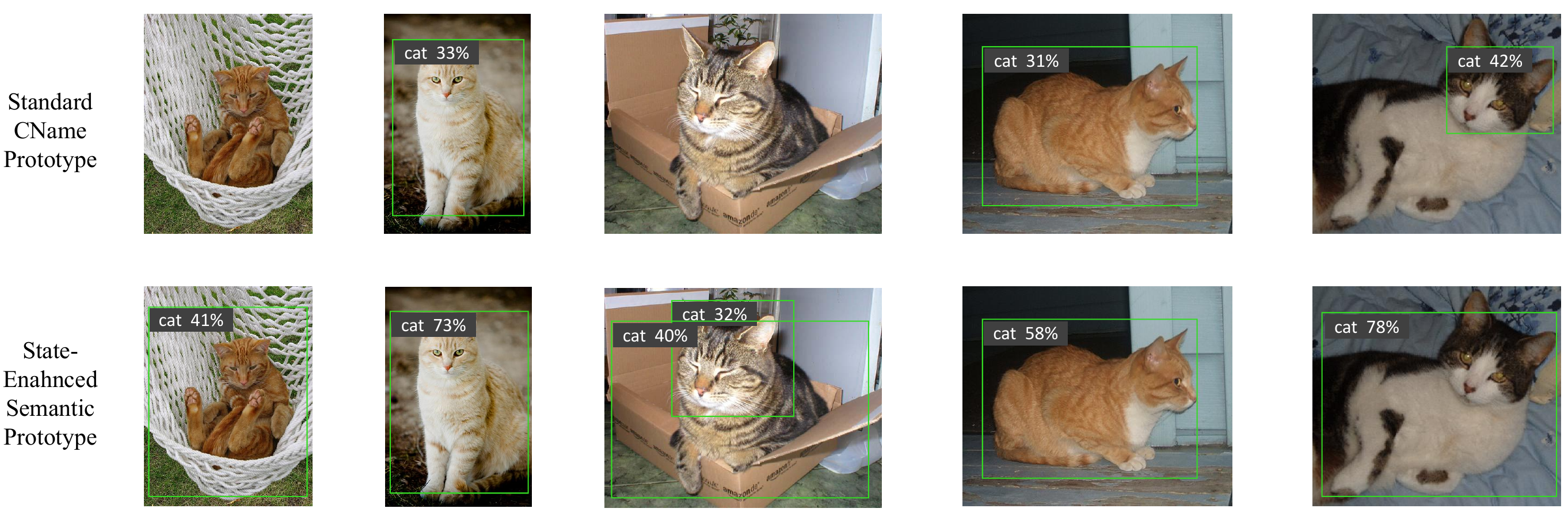}
   \caption{Qualitative comparison of detection results using standard class-name (CName) prototypes (top row) versus our state-enhanced semantic prototypes (bottom row). The detector is trained on the OV-COCO dataset. At inference time, only the ``cat'' category is treated as the target class, and predictions are filtered with a confidence threshold of 0.3.}
   \label{fig:qual}
\end{figure*}

\subsection{Cross-Dataset Evaluation}
To further investigate the generalization of the proposed method, we conduct cross-dataset transfer evaluation in \cref{tab:objects365}.  
We train the detector on OV-LVIS and evaluate on the validation set of Objects365~\cite{shao2019objects365}. Following \cite{kaul2023multi}, all models are trained on the full LVIS training set with a ResNet-50 backbone, leveraging ImageNet images labeled with LVIS categories as extra weak supervision. During inference, we replace the LVIS-based text classifiers with new classifiers built for Objects365: for each Objects365 category, we generate generic and state-specific textual descriptions and aggregate their CLIP embeddings to construct state-enhanced semantic prototypes, while keeping all other model parameters unchanged. In Objects365, we define the ``rare'' subset as the one-third of categories with the lowest instance frequency in the training set, following \cite{kaul2023multi}, and report performance on this subset using the AP$_r$ metric.

As shown in \cref{tab:objects365}, our approach consistently outperforms Detic~\cite{zhou2022detecting} and MM-OVOD-Text~\cite{kaul2023multi} models under the same cross-dataset setting across all metrics. These results demonstrate that enriching class representations with state-aware descriptions—rather than relying solely on class names or generic textual prompts—yields more transferable semantic knowledge, even in the challenging cross-dataset scenario.

\subsection{Qualitative Detection Comparison}
To evaluate the generalization capability of our state-enhanced semantic prototypes across diverse object appearances, we visualize detection results on images depicting cats in various poses, including sitting, lying, stretching, and partial occlusion, in \cref{fig:qual}. Standard class name prototypes often struggle with pose variation, producing low-confidence detections, especially when the object appears in atypical poses or is partially visible. In contrast, our state-enhanced semantic prototypes generate higher-confidence predictions and more accurate localizations, even under challenging deformations or occlusions. This improvement demonstrates that by integrating state-aware semantic information into prototype construction, our method effectively captures the rich intra-class visual variations commonly observed in real-world object instances. Consequently, the enhanced semantic alignment between language and vision enables more reliable open-vocabulary object detection, particularly for categories exhibiting significant appearance diversity.

\section{Conclusion}
In this work, we address two key limitations in weakly supervised open-vocabulary object detection (WS-OVOD): the inability of standard semantic prototypes to capture intra-class visual variations caused by diverse object states, and the semantic mismatch between visual proposals and text embeddings under weak supervision. To this end, we propose a dual-augmentation framework that jointly enhances visual-textual alignment from two complementary perspectives. First, we introduce state-enhanced semantic prototypes (SESP) by leveraging large language models to generate diverse, state-aware textual descriptions, capturing intra-class visual variations such as pose or activity. Second, we propose scene-augmented pseudo prototypes (SAPP) that encode object-context co-occurrence cues (e.g., ``cat on sofa'') and align them softly with weakly supervised max-size proposals to bridge the contextual gap between region features and semantic representations. By integrating SESP and SAPP, our method achieves more discriminative and contextually grounded representations, leading to consistent improvements in novel category detection under weak supervision. Experimental results on multiple benchmarks demonstrate the effectiveness of our proposed method.
{
    \small
    \nocite{*}
    \bibliographystyle{ieeenat_fullname}
    \bibliography{main}

@String(PAMI = {IEEE Trans. Pattern Anal. Mach. Intell.})

@String(CVPR= {IEEE Conf. Comput. Vis. Pattern Recog.})

@String(AAAI = {AAAI})

@String(PAMI  = {IEEE TPAMI})

@String(CVPR  = {CVPR})

@misc{Authors14,
 author = {FirstName LastName},
 title = {The frobnicatable foo filter},
 note = {Face and Gesture submission ID 324. Supplied as supplemental material {\tt fg324.pdf}},
 year = 2014
}

@misc{Authors14b,
 author = {FirstName LastName},
 title = {Frobnication tutorial},
 note = {Supplied as supplemental material {\tt tr.pdf}},
 year = 2014
}

@article{Alpher02,
author = {FirstName Alpher},
title = {Frobnication},
journal = PAMI,
volume = 12,
number = 1,
pages = {234--778},
year = 2002
}

@article{Alpher03,
author = {FirstName Alpher and  FirstName Fotheringham-Smythe},
title = {Frobnication revisited},
journal = {Journal of Foo},
volume = 13,
number = 1,
pages = {234--778},
year = 2003
}

@article{Alpher04,
author = {FirstName Alpher and FirstName Fotheringham-Smythe and FirstName Gamow},
title = {Can a machine frobnicate?},
journal = {Journal of Foo},
volume = 14,
number = 1,
pages = {234--778},
year = 2004
}

@inproceedings{Alpher05,
author = {FirstName Alpher and FirstName Gamow},
title = {Can a computer frobnicate?},
booktitle = CVPR,
pages = {234--778},
year = 2005
}

@inproceedings{OVR-CNN,
  title={Open-vocabulary object detection using captions},
  author={Zareian, Alireza and Rosa, Kevin Dela and Hu, Derek Hao and Chang, Shih-Fu},
  booktitle={Proceedings of the IEEE/CVF conference on computer vision and pattern recognition},
  pages={14393--14402},
  year={2021}
}

@inproceedings{glip,
  title={Grounded language-image pre-training},
  author={Li, Liunian Harold and Zhang, Pengchuan and Zhang, Haotian and Yang, Jianwei and Li, Chunyuan and Zhong, Yiwu and Wang, Lijuan and Yuan, Lu and Zhang, Lei and Hwang, Jenq-Neng and others},
  booktitle={Proceedings of the IEEE/CVF conference on computer vision and pattern recognition},
  pages={10965--10975},
  year={2022}
}

@inproceedings{rovit,
  title={Region-aware pretraining for open-vocabulary object detection with vision transformers},
  author={Kim, Dahun and Angelova, Anelia and Kuo, Weicheng},
  booktitle={Proceedings of the IEEE/CVF conference on computer vision and pattern recognition},
  pages={11144--11154},
  year={2023}
}

@article{lin2022learning,
  title={Learning object-language alignments for open-vocabulary object detection},
  author={Lin, Chuang and Sun, Peize and Jiang, Yi and Luo, Ping and Qu, Lizhen and Haffari, Gholamreza and Yuan, Zehuan and Cai, Jianfei},
  journal={arXiv preprint arXiv:2211.14843},
  year={2022}
}

@inproceedings{zhong2022regionclip,
  title={Regionclip: Region-based language-image pretraining},
  author={Zhong, Yiwu and Yang, Jianwei and Zhang, Pengchuan and Li, Chunyuan and Codella, Noel and Li, Liunian Harold and Zhou, Luowei and Dai, Xiyang and Yuan, Lu and Li, Yin and others},
  booktitle={Proceedings of the IEEE/CVF conference on computer vision and pattern recognition},
  pages={16793--16803},
  year={2022}
}

@inproceedings{fu2025llmdet,
  title={Llmdet: Learning strong open-vocabulary object detectors under the supervision of large language models},
  author={Fu, Shenghao and Yang, Qize and Mo, Qijie and Yan, Junkai and Wei, Xihan and Meng, Jingke and Xie, Xiaohua and Zheng, Wei-Shi},
  booktitle={Proceedings of the Computer Vision and Pattern Recognition Conference},
  pages={14987--14997},
  year={2025}
}

@inproceedings{liu2024groundingdino,
  title={Grounding dino: Marrying dino with grounded pre-training for open-set object detection},
  author={Liu, Shilong and Zeng, Zhaoyang and Ren, Tianhe and Li, Feng and Zhang, Hao and Yang, Jie and Jiang, Qing and Li, Chunyuan and Yang, Jianwei and Su, Hang and others},
  booktitle={European conference on computer vision},
  pages={38--55},
  year={2024},
  organization={Springer}
}

@article{gu2021open,
  title={Open-vocabulary object detection via vision and language knowledge distillation},
  author={Gu, Xiuye and Lin, Tsung-Yi and Kuo, Weicheng and Cui, Yin},
  journal={arXiv preprint arXiv:2104.13921},
  year={2021}
}

@inproceedings{clip,
  title={Learning transferable visual models from natural language supervision},
  author={Radford, Alec and Kim, Jong Wook and Hallacy, Chris and Ramesh, Aditya and Goh, Gabriel and Agarwal, Sandhini and Sastry, Girish and Askell, Amanda and Mishkin, Pamela and Clark, Jack and others},
  booktitle={International conference on machine learning},
  pages={8748--8763},
  year={2021},
  organization={PmLR}
}

@inproceedings{wu2023aligning,
  title={Aligning bag of regions for open-vocabulary object detection},
  author={Wu, Size and Zhang, Wenwei and Jin, Sheng and Liu, Wentao and Loy, Chen Change},
  booktitle={Proceedings of the IEEE/CVF conference on computer vision and pattern recognition},
  pages={15254--15264},
  year={2023}
}

@inproceedings{ma2022open,
  title={Open-vocabulary one-stage detection with hierarchical visual-language knowledge distillation},
  author={Ma, Zongyang and Luo, Guan and Gao, Jin and Li, Liang and Chen, Yuxin and Wang, Shaoru and Zhang, Congxuan and Hu, Weiming},
  booktitle={Proceedings of the IEEE/CVF conference on computer vision and pattern recognition},
  pages={14074--14083},
  year={2022}
}

@inproceedings{oadp,
  title={Object-aware distillation pyramid for open-vocabulary object detection},
  author={Wang, Luting and Liu, Yi and Du, Penghui and Ding, Zihan and Liao, Yue and Qi, Qiaosong and Chen, Biaolong and Liu, Si},
  booktitle={Proceedings of the IEEE/CVF Conference on Computer Vision and Pattern Recognition},
  pages={11186--11196},
  year={2023}
}

@article{hpovd,
  title={A Hierarchical Semantic Distillation Framework for Open-Vocabulary Object Detection},
  author={Fu, Shenghao and Yan, Junkai and Yang, Qize and Wei, Xihan and Xie, Xiaohua and Zheng, Wei-Shi},
  journal={arXiv preprint arXiv:2503.10152},
  year={2025}
}

@inproceedings{dkdetr,
  title={Distilling detr with visual-linguistic knowledge for open-vocabulary object detection},
  author={Li, Liangqi and Miao, Jiaxu and Shi, Dahu and Tan, Wenming and Ren, Ye and Yang, Yi and Pu, Shiliang},
  booktitle={Proceedings of the IEEE/CVF International Conference on Computer Vision},
  pages={6501--6510},
  year={2023}
}

@inproceedings{ma2025cake,
  title={Cake: Category aware knowledge extraction for open-vocabulary object detection},
  author={Ma, Shiyuan and Qian, Donglin and Ye, Kai and Zhang, Shengchuan},
  booktitle={Proceedings of the AAAI Conference on Artificial Intelligence},
  volume={39},
  number={6},
  pages={5982--5990},
  year={2025}
}

@inproceedings{kaul2023multi,
  title={Multi-modal classifiers for open-vocabulary object detection},
  author={Kaul, Prannay and Xie, Weidi and Zisserman, Andrew},
  booktitle={International Conference on Machine Learning},
  pages={15946--15969},
  year={2023},
  organization={PMLR}
}

@article{jin2024llms,
  title={Llms meet vlms: Boost open vocabulary object detection with fine-grained descriptors},
  author={Jin, Sheng and Jiang, Xueying and Huang, Jiaxing and Lu, Lewei and Lu, Shijian},
  journal={arXiv preprint arXiv:2402.04630},
  year={2024}
}

@inproceedings{liu2024shine,
  title={Shine: Semantic hierarchy nexus for open-vocabulary object detection},
  author={Liu, Mingxuan and Hayes, Tyler L and Ricci, Elisa and Csurka, Gabriela and Volpi, Riccardo},
  booktitle={Proceedings of the IEEE/CVF Conference on Computer Vision and Pattern Recognition},
  pages={16634--16644},
  year={2024}
}

@inproceedings{du2022learning,
  title={Learning to prompt for open-vocabulary object detection with vision-language model},
  author={Du, Yu and Wei, Fangyun and Zhang, Zihe and Shi, Miaojing and Gao, Yue and Li, Guoqi},
  booktitle={Proceedings of the IEEE/CVF conference on computer vision and pattern recognition},
  pages={14084--14093},
  year={2022}
}

@inproceedings{feng2022promptdet,
  title={Promptdet: Towards open-vocabulary detection using uncurated images},
  author={Feng, Chengjian and Zhong, Yujie and Jie, Zequn and Chu, Xiangxiang and Ren, Haibing and Wei, Xiaolin and Xie, Weidi and Ma, Lin},
  booktitle={European conference on computer vision},
  pages={701--717},
  year={2022},
  organization={Springer}
}

@article{ren2023prompt,
  title={Prompt pre-training with twenty-thousand classes for open-vocabulary visual recognition},
  author={Ren, Shuhuai and Zhang, Aston and Zhu, Yi and Zhang, Shuai and Zheng, Shuai and Li, Mu and Smola, Alexander J and Sun, Xu},
  journal={Advances in Neural Information Processing Systems},
  volume={36},
  pages={12569--12588},
  year={2023}
}

@inproceedings{zhao2024scene,
  title={Scene-adaptive and region-aware multi-modal prompt for open vocabulary object detection},
  author={Zhao, Xiaowei and Liu, Xianglong and Wang, Duorui and Gao, Yajun and Liu, Zhide},
  booktitle={Proceedings of the IEEE/CVF Conference on Computer Vision and Pattern Recognition},
  pages={16741--16750},
  year={2024}
}

@article{bangalath2022bridging,
  title={Bridging the gap between object and image-level representations for open-vocabulary detection},
  author={Bangalath, Hanoona and Maaz, Muhammad and Khattak, Muhammad Uzair and Khan, Salman H and Shahbaz Khan, Fahad},
  journal={Advances in Neural Information Processing Systems},
  volume={35},
  pages={33781--33794},
  year={2022}
}

@inproceedings{gao2022open,
  title={Open vocabulary object detection with pseudo bounding-box labels},
  author={Gao, Mingfei and Xing, Chen and Niebles, Juan Carlos and Li, Junnan and Xu, Ran and Liu, Wenhao and Xiong, Caiming},
  booktitle={European Conference on Computer Vision},
  pages={266--282},
  year={2022},
  organization={Springer}
}

@inproceedings{zhou2022detecting,
  title={Detecting twenty-thousand classes using image-level supervision},
  author={Zhou, Xingyi and Girdhar, Rohit and Joulin, Armand and Kr{\"a}henb{\"u}hl, Philipp and Misra, Ishan},
  booktitle={European conference on computer vision},
  pages={350--368},
  year={2022},
  organization={Springer}
}

@inproceedings{jeong2024proxydet,
  title={Proxydet: Synthesizing proxy novel classes via classwise mixup for open-vocabulary object detection},
  author={Jeong, Joonhyun and Park, Geondo and Yoo, Jayeon and Jung, Hyungsik and Kim, Heesu},
  booktitle={Proceedings of the AAAI Conference on Artificial Intelligence},
  volume={38},
  number={3},
  pages={2462--2470},
  year={2024}
}

@article{huang2024open,
  title={Open-vocabulary object detection via language hierarchy},
  author={Huang, Jiaxing and Zhang, Jingyi and Jiang, Kai and Lu, Shijian},
  journal={arXiv preprint arXiv:2410.20371},
  year={2024}
}

@article{ma2023codet,
  title={Codet: Co-occurrence guided region-word alignment for open-vocabulary object detection},
  author={Ma, Chuofan and Jiang, Yi and Wen, Xin and Yuan, Zehuan and Qi, Xiaojuan},
  journal={Advances in neural information processing systems},
  volume={36},
  pages={71078--71094},
  year={2023}
}

@inproceedings{pham2024lp,
  title={LP-OVOD: Open-vocabulary object detection by linear probing},
  author={Pham, Chau and Vu, Truong and Nguyen, Khoi},
  booktitle={Proceedings of the IEEE/CVF Winter Conference on Applications of Computer Vision},
  pages={779--788},
  year={2024}
}

@inproceedings{zhao2024taming,
  title={Taming self-training for open-vocabulary object detection},
  author={Zhao, Shiyu and Schulter, Samuel and Zhao, Long and Zhang, Zhixing and Suh, Yumin and Chandraker, Manmohan and Metaxas, Dimitris N and others},
  booktitle={Proceedings of the IEEE/CVF Conference on Computer Vision and Pattern Recognition},
  pages={13938--13947},
  year={2024}
}

@article{xu2023dst,
  title={Dst-det: Simple dynamic self-training for open-vocabulary object detection},
  author={Xu, Shilin and Li, Xiangtai and Wu, Size and Zhang, Wenwei and Tong, Yunhai and Loy, Chen Change},
  journal={arXiv preprint arXiv:2310.01393},
  year={2023}
}

@inproceedings{bilen2016weakly,
  title={Weakly supervised deep detection networks},
  author={Bilen, Hakan and Vedaldi, Andrea},
  booktitle={Proceedings of the IEEE conference on computer vision and pattern recognition},
  pages={2846--2854},
  year={2016}
}

@inproceedings{ren2020instance,
  title={Instance-aware, context-focused, and memory-efficient weakly supervised object detection},
  author={Ren, Zhongzheng and Yu, Zhiding and Yang, Xiaodong and Liu, Ming-Yu and Lee, Yong Jae and Schwing, Alexander G and Kautz, Jan},
  booktitle={Proceedings of the IEEE/CVF conference on computer vision and pattern recognition},
  pages={10598--10607},
  year={2020}
}

@inproceedings{tang2017multiple,
  title={Multiple instance detection network with online instance classifier refinement},
  author={Tang, Peng and Wang, Xinggang and Bai, Xiang and Liu, Wenyu},
  booktitle={Proceedings of the IEEE conference on computer vision and pattern recognition},
  pages={2843--2851},
  year={2017}
}

@inproceedings{li2019weakly,
  title={Weakly supervised object detection with segmentation collaboration},
  author={Li, Xiaoyan and Kan, Meina and Shan, Shiguang and Chen, Xilin},
  booktitle={Proceedings of the IEEE/CVF international conference on computer vision},
  pages={9735--9744},
  year={2019}
}

@inproceedings{wan2019c,
  title={C-mil: Continuation multiple instance learning for weakly supervised object detection},
  author={Wan, Fang and Liu, Chang and Ke, Wei and Ji, Xiangyang and Jiao, Jianbin and Ye, Qixiang},
  booktitle={Proceedings of the IEEE/CVF conference on computer vision and pattern recognition},
  pages={2199--2208},
  year={2019}
}

@inproceedings{diba2017weakly,
  title={Weakly supervised cascaded convolutional networks},
  author={Diba, Ali and Sharma, Vivek and Pazandeh, Ali and Pirsiavash, Hamed and Van Gool, Luc},
  booktitle={Proceedings of the IEEE conference on computer vision and pattern recognition},
  pages={914--922},
  year={2017}
}

@inproceedings{zhang2018adversarial,
  title={Adversarial complementary learning for weakly supervised object localization},
  author={Zhang, Xiaolin and Wei, Yunchao and Feng, Jiashi and Yang, Yi and Huang, Thomas S},
  booktitle={Proceedings of the IEEE conference on computer vision and pattern recognition},
  pages={1325--1334},
  year={2018}
}

@inproceedings{redmon2017yolo9000,
  title={YOLO9000: better, faster, stronger},
  author={Redmon, Joseph and Farhadi, Ali},
  booktitle={Proceedings of the IEEE conference on computer vision and pattern recognition},
  pages={7263--7271},
  year={2017}
}

@inproceedings{ramanathan2020dlwl,
  title={Dlwl: Improving detection for lowshot classes with weakly labelled data},
  author={Ramanathan, Vignesh and Wang, Rui and Mahajan, Dhruv},
  booktitle={Proceedings of the IEEE/CVF conference on computer vision and pattern recognition},
  pages={9342--9352},
  year={2020}
}

@article{shao2022deep,
  title={Deep learning for weakly-supervised object detection and localization: A survey},
  author={Shao, Feifei and Chen, Long and Shao, Jian and Ji, Wei and Xiao, Shaoning and Ye, Lu and Zhuang, Yueting and Xiao, Jun},
  journal={Neurocomputing},
  volume={496},
  pages={192--207},
  year={2022},
  publisher={Elsevier}
}

@inproceedings{lin2014microsoft,
  title={Microsoft coco: Common objects in context},
  author={Lin, Tsung-Yi and Maire, Michael and Belongie, Serge and Hays, James and Perona, Pietro and Ramanan, Deva and Doll{\'a}r, Piotr and Zitnick, C Lawrence},
  booktitle={European conference on computer vision},
  pages={740--755},
  year={2014},
  organization={Springer}
}

@inproceedings{gupta2019lvis,
  title={Lvis: A dataset for large vocabulary instance segmentation},
  author={Gupta, Agrim and Dollar, Piotr and Girshick, Ross},
  booktitle={Proceedings of the IEEE/CVF conference on computer vision and pattern recognition},
  pages={5356--5364},
  year={2019}
}

@article{ren2016fasterrcnn,
  title={Faster R-CNN: Towards real-time object detection with region proposal networks},
  author={Ren, Shaoqing and He, Kaiming and Girshick, Ross and Sun, Jian},
  journal={IEEE transactions on pattern analysis and machine intelligence},
  volume={39},
  number={6},
  pages={1137--1149},
  year={2016},
  publisher={IEEE}
}

@inproceedings{he2016deepresnet,
  title={Deep residual learning for image recognition},
  author={He, Kaiming and Zhang, Xiangyu and Ren, Shaoqing and Sun, Jian},
  booktitle={Proceedings of the IEEE conference on computer vision and pattern recognition},
  pages={770--778},
  year={2016}
}

@article{zhou2021probabilisticcenternet2,
  title={Probabilistic two-stage detection},
  author={Zhou, Xingyi and Koltun, Vladlen and Kr{\"a}henb{\"u}hl, Philipp},
  journal={arXiv preprint arXiv:2103.07461},
  year={2021}
}

@article{achiam2023gpt,
  title={Gpt-4 technical report},
  author={Achiam, Josh and Adler, Steven and Agarwal, Sandhini and Ahmad, Lama and Akkaya, Ilge and Aleman, Florencia Leoni and Almeida, Diogo and Altenschmidt, Janko and Altman, Sam and Anadkat, Shyamal and others},
  journal={arXiv preprint arXiv:2303.08774},
  year={2023}
}

@inproceedings{li2025benefit,
  title={Benefit From Seen: Enhancing Open-Vocabulary Object Detection by Bridging Visual and Textual Co-Occurrence Knowledge},
  author={Li, Yanqi and Niu, Jianwei and Ren, Tao},
  booktitle={Proceedings of the IEEE/CVF International Conference on Computer Vision},
  pages={22110--22119},
  year={2025}
}

@article{zhang2024opendcs,
  title={Open-vocabulary object detection via debiased curriculum self-training},
  author={Zhang, Hanlue and Guan, Dayan and Ke, Xiangrui and El Saddik, Abdulmotaleb and Lu, Shijian},
  journal={Expert Systems with Applications},
  volume={255},
  pages={124762},
  year={2024},
  publisher={Elsevier}
}

@inproceedings{shao2019objects365,
  title={Objects365: A large-scale, high-quality dataset for object detection},
  author={Shao, Shuai and Li, Zeming and Zhang, Tianyuan and Peng, Chao and Yu, Gang and Zhang, Xiangyu and Li, Jing and Sun, Jian},
  booktitle={Proceedings of the IEEE/CVF international conference on computer vision},
  pages={8430--8439},
  year={2019}
}

@inproceedings{wang2023opencorpus,
  title={Open-vocabulary object detection with an open corpus},
  author={Wang, Jiong and Zhang, Huiming and Hong, Haiwen and Jin, Xuan and He, Yuan and Xue, Hui and Zhao, Zhou},
  booktitle={Proceedings of the IEEE/CVF international conference on computer vision},
  pages={6759--6769},
  year={2023}
}

@inproceedings{deng2009imagenet,
  title={Imagenet: A large-scale hierarchical image database},
  author={Deng, Jia and Dong, Wei and Socher, Richard and Li, Li-Jia and Li, Kai and Fei-Fei, Li},
  booktitle={2009 IEEE conference on computer vision and pattern recognition},
  pages={248--255},
  year={2009},
  organization={Ieee}
}

@inproceedings{sharma2018conceptual,
  title={Conceptual captions: A cleaned, hypernymed, image alt-text dataset for automatic image captioning},
  author={Sharma, Piyush and Ding, Nan and Goodman, Sebastian and Soricut, Radu},
  booktitle={Proceedings of the 56th Annual Meeting of the Association for Computational Linguistics (Volume 1: Long Papers)},
  pages={2556--2565},
  year={2018}
}

@inproceedings{fang2024simple,
  title={Simple image-level classification improves open-vocabulary object detection},
  author={Fang, Ruohuan and Pang, Guansong and Bai, Xiao},
  booktitle={Proceedings of the AAAI conference on artificial intelligence},
  volume={38},
  number={2},
  pages={1716--1725},
  year={2024}
}

@article{ma2025hierarchical,
  title={Hierarchical Multimodal Knowledge Matching for Training-Free Open-Vocabulary Object Detection},
  author={Ma, Qisen and Huang, Yan and Liu, Zikun and Park, Hyunhee and Wang, Liang},
  journal={IEEE Transactions on Image Processing},
  year={2025},
  publisher={IEEE}
}
}



\end{document}